\documentclass[%
 reprint,
 amsmath,amssymb,
 aps, 
]{revtex4-2}
\usepackage{xcolor}
\usepackage{graphicx}
\usepackage{dcolumn}
\usepackage{bm}

\begin{document}

\preprint{APS/123-QED}

\title{Bayesian Deep Learning Hyperparameter Search for Robust \\Function Mapping to Polynomials with Noise}

\author{Nidhin Harilal}
\affiliation{%
Indian Institute of Technology Gandhinagar, Gujarat, India \\nidhin.harilal@iitgn.ac.in
}%

\author{Udit Bhatia}
\affiliation{
Indian Institute of Technology Gandhinagar, Gujarat, India \\bhatia.u@iitgn.ac.in
}%

\author{Auroop R. Ganguly}
\affiliation{
Northeastern University, Boston, MA, USA\\ Pacific Northwest National Laboratory, Richland, WA, USA\\a.ganguly@northeastern.edu
}%


\date{\today}

\begin{abstract}

Advances in neural architecture search, as well as explainability and interpretability of connectionist architectures, have been reported in the recent literature. However, our understanding of how to design Bayesian Deep Learning (BDL) hyperparameters, specifically, the depth, width and ensemble size, for robust function mapping with uncertainty quantification, is still emerging. This paper attempts to further our understanding by mapping Bayesian connectionist representations to polynomials of different orders with varying noise types and ratios. We examine the noise-contaminated polynomials to search for the combination of hyperparameters that can extract the underlying polynomial signals while quantifying uncertainties based on the noise attributes. Specifically, we attempt to study the question that an appropriate neural architecture and ensemble configuration can be found to detect a signal of any n-th (where n $\in$ N) order polynomial contaminated with noise having different distributions and signal-to-noise (SNR) ratios and varying noise attributes. Our results suggest the possible existence of an optimal network depth as well as an optimal number of ensembles for prediction skills and uncertainty quantification, respectively. However, optimality is not discernible for width, even though the performance gain reduces with increasing width at high values of width. Our experiments and insights can be directional to understand theoretical properties of BDL representations and to design practical solutions.

\end{abstract}
\maketitle

\newpage
\section{\label{sec:intro}Introduction}
Neural Networks (NNs) or connectionist representations were originally inspired by the human brain~\cite{amit1992modeling}, while feedforward NNs or MultiLayer Perceptrons (MLPs) were later shown to act as universal function approximators~\cite{hornik1989multilayer, hornik1991approximation,  barron1993universal}. However, recent literature points to the imperfect nature of biological analogies for NNs~\cite{lillicrap2020backpropagation} and the “unreasonable effectiveness” of deep learning~\cite{sejnowski2020unreasonable}, or deep NN representations. Bayesian methods for uncertainty quantification (UQ) have been suggested for both shallow~\cite{mackay1995bayesian, neal1993bayesian, neal2012bayesian, kendall2017uncertainties} and deep~\cite{gal2016dropout, gal2017concrete, vandal2018quantifying} NNs. Despite recent successes of connectionist architectures~\cite{norris1994shortlist, christiansen1999toward}, especially deep learning NNs~\cite{krizhevsky2012imagenet, lecun2015deep, sejnowski2020unreasonable} including Bayesian Deep Learning (BDL)~\cite{kendall2017uncertainties, khan2018fast}, major gaps remain in our theoretical understanding and in the design of practical solutions. Deep learning representations, in particular, appear at first glance to defy the principle of Occam's Razor or model parsimony, even though \textcolor{black}{Bayesian~\cite{gal2016dropout, kendall2017uncertainties, vasudevan2021off, luo2020bayesian}} or \textcolor{black}{physics-guided~\cite{jia2021physics, jia2019physics, li2018exploring}} approaches may be viewed as constraining the plausible hypotheses space. Given the simplifying assumptions often made to establish NN (including deep learning) theory and the ad hoc nature of most engineering solutions, a complementary approach may be rigorous design-of-experiments with simulated data. Here we design a set of experiments to understand the function approximation capability of NNs including deep representations. We map a set of NNs, specifically (shallow and deep) MLPs, to a set of polynomials contaminated with noise. The mapping is explored keeping in mind that both NNs and polynomials are universal function approximators (UFAs) in principle. We simulate data by varying the degree of the polynomials along with the type of noise and the signal-to-noise ratios (SNR). The hyperparameters of the NN (i.e., MLP) representations (i.e., depth and width for the connectionist function representations and ensemble size for UQ) are examined to characterize the robustness of the fit in terms of the ability to recover the original polynomial (measured through prediction skill on test data) and the noise attributes (measured through distributional distance metrics).     

Cybenko~\cite{cybenko1989approximation}, Funhashi~\cite{funahashi1989approximate} and Hornik et al.~\cite{hornik1989multilayer} proved that a finite linear sum of continuous sigmoidal functions could approximate any function to a desired degree of accuracy, or in other words, that MLPs act as UFAs. Subsequently, numerous results have shown the UFA property of NNs for different function \textcolor{black}{classes~\cite{hornik1991approximation, park1991universal, barron1993universal, leshno1993multilayer, bengio2011expressive}}. Mathematically, this universal approximation property of neural networks can be described as: Given a continuous target goal function $f(x)$, there exists a output function $g(x)$ in a linear summation form: 
\begin{eqnarray}
g(x) = \Sigma_{i=1}^{n_h}w_{2,i}\sigma(W_{1,i}x + b_i) \\
\text{where}\; \sigma(x) = \frac{1}{1+e^{-x}}
\end{eqnarray}

and $w_{1,i}$, $w_{2,i}$ are the weights between the first and the output layers respectively and $b_i$ are the bias values associated with the layer such that $|g(x) - f(x)| \le \epsilon$ for all $x$ where $\epsilon$ is an arbitrarily small number. 

The UFA property of NNs show that, under various regularity assumptions and given sufficient data, they can approximate a function $f(t)$ to any desired degree of accuracy. This implies that for an adequate number of hidden layers ($l$) and nodes ($n_h$), there exists a set of corresponding weights and biases to achieve function approximation to any desired level of accuracy. However, the theory does not prescribe what the $l$ and $n_h$ should be, thus imposing a significant practical challenge. Also, the non-unique trained parameters of such NNs are difficult to interpret or explain. Given the growing complexity of deep learning models and the associated computational challenges, model parsimony and neural architecture search have become crucial research areas. 

{The possible existence of an optimal depth in NNs has been explored in NN hyperparameter search. This body of literature~\cite{eldan2016power, cohen2016expressive, liang2016deep} shows the existence of certain deep ReLU networks that cannot be realized form shallow networks. Zhou et al.~\cite{lu2017expressive} on the other hand, explores how width affects the expressiveness of neural networks and shows the existence of classes of wide networks which cannot be realized by any narrow network whose depth is no more than a polynomial bound. Poggio et al.~\cite{poggio2017and} provide theorems and examples of a class of compositional functions for which there is a gap between approximation in shallow and deep networks.}

{Theoretical papers on ANNs including MLPs and DL, have focused on performance guarantees often based on simplified \textcolor{black}{assumptions~\cite{eldan2016power, cohen2016expressive, liang2016deep, bengio2011expressive}}. 
However, what has received limited attention in the literature is the influence of hyperparameter selection on robust out-of-sample generalization and uncertainty characterization. One of the key challenges in the analysis of  generalization bounds in deep neural networks is that it may vary depending on the data distribution on which networks is trained. Therefore, such an exploration requires extensive empirical analysis. It is however, not practically possible to experiment on all possible distributions that may exist. Therefore, the analysis needs to be restricted to certain family of distributions. Itay et al.~\cite{safran2017depth} provides a theoritical and empirical analysis to provide several new depth-based separation results on natural radial non-linear functions such as balls and ellipses. However, their empirical analysis is restricted to just two depth values.}

Recent attempts at UQ on NNs rely on what have been called Bayesian approaches~\cite{izmailov2020subspace} and have taken the form of so-called  \textcolor{black}{BDL~\cite{wang2016towards, gal2016dropout, luo2020bayesian}}. One way of incorporating Bayesian inferencing in neural networks relies on ensembles developed through random selection of nodes via a “dropout” strategy (MC-dropout based networks)~\cite{gal2016dropout}. The complexity of conventional neural network representations for point predictions, along with the heuristic nature of BDL for uncertainty quantification, implies that any specific DL (or BDL) based function approximation need to be carefully examined by the ability to delineate and distinguish between what may be viewed as the signal (with repeatable or generalizable patterns that may be deterministic) versus noise 
(which is not repeatable, indeed it is usually modeled as a stochastic process). 

In this paper, we examine the ability of MC-dropout based neural networks to distinguish between signal and noise in polynomials of different orders contaminated with different levels and types of uncorrelated random samples. Our underlying hypothesis is that the DL-based point prediction can capture the underlying order of the polynomial while the BDL-based uncertainty quantification can delineate and capture the statistical attributes of the noise. A second associated hypothesis - based on the UFA properties of both NNs and polynomials - is that the behavior exhibited by polynomials contaminated with noise can be expressed through NNs (specifically, MLPs) with different hyperparameters. The simulation-based experimental design examine the hypotheses via metrics for skills in prediction and UQ.

\section{\label{sec:exp}Experimental Design}
In this section, we discuss the overview of the experiments performed to explore the performance of BDL in terms of modeling the underlying polynomial and approximating noise attributes from data sets of noise-contaminated polynomials.

\subsection{Polynomial Dataset}

We consider polynomials of different orders and coefficients and add different types of noises with varied SNR with an understanding that polynomials are Universal Function Approximators (UFAs) themselves ~\cite{pinkus2000weierstrass}. We use 3 types of noises: (a) Gaussian, (b) Exponential, ad (c) Rayleigh with SNR ranging from 10 to 30.  Mathematically, this polynomial dataset ($P$) can be represented as follows in Eq. (\ref{eq:polynomial}):
\begin{eqnarray}
\label{eq:polynomial}
P = p(x, n) + \epsilon_{D(t, r)} \;\;\text{where, }\, p(x,n) = \Sigma_{i=0}^{n} a_{i}x^{i}
\end{eqnarray}

where $\epsilon_{D(t, snr)}$ represents the noise from distribution $t$ with SNR level $r$. Table~\ref{data_stat} shows the details of various attributes of attributed dataset.

\begin{table}[]
\caption{%
Polynomial dataset description
}
\begin{ruledtabular}
\begin{tabular}{cc}
\label{data_stat}
\textbf{Attribute} & \textbf{Types}\\
\hline
Poly. Order ($n$) & 2, 3, 4, 5, 7, 10 \\
Noise Types ($t$) &  Gaussian\footnote{$f(x; \mu; \sigma) = \frac{1}{\sigma \sqrt{2\pi}} e^{-(x-\mu)^{2} / (2\sigma^2)}$}, 
Exponential\footnote{$
  f(x; \lambda) =
    \begin{cases}
      \lambda e^{-\lambda x} & x\ge 0\\
      0 & \text{otherwise}
    \end{cases}$
}, 
Rayleigh\footnote{$f(x; \sigma) = \frac{x}{\sigma^2}e^{-x^{2}* (2\sigma^2)}$} \\
SNR Levels ($r$) & 10, 20 30\\ 
\end{tabular}
\end{ruledtabular}
\end{table}

\subsection{Neural Network Design}
To understand how different levels of abstraction in neural networks affect their representation power, we vary the following attributes in our experiments: (a) number of hidden layers; (b) nodes in each layer; and (c) number of ensembles.  

We follow a bottom-up approach for designing the network configuration for our experiments. That is, we start with a 1-layer neural network with six neurons and gradually increase the width and depth of the network and each time we initialize the network with random weights. Consider width $w \in \mathbb{N}$, depth $d \in \mathbb{N}$ and number of ensembles $m \in \mathbb{N}$, we represent $G(w, d, m)$ for a feedforward neural network with $d$-layers each having $w$-neurons and the network having $m$-ensembles. Table~\ref{tab:poly_data} shows the the considered set of values of different hyperparamters. We analyze the behaviour of $G(w,d,m)$ on our synthetically generated noisy-polynomial data. This is done by training each candidate network $G(w,d,m)$ on polynomial data with different order and noise levels. Details regarding these polynomial orders and noise types are given in Table~\ref{data_stat}. This experiment is repeated until all possible hyperparamter combinations have been trained on all synthetically generated data. 

\begin{table}[hb]
\caption{\label{tab:poly_data}%
Network Configurations
}
\begin{ruledtabular}
\begin{tabular}{cc}
\textbf{Parameter} & \textbf{Types}\\
\hline
Depth & 1, 2, 3, 4, ..., 15 \\
Width & 6, 8, 10 16, 20, 30, ..., 1000\\
No. of Ensembles & 1, 5, 10, 20, 30, ...,40\\ 
\end{tabular}
\end{ruledtabular}
\end{table}

\subsection{Evaluation}
This section discusses different criteria used for evaluating the degree of approximation and robustness of the trained networks.

\textbf{L2-norm:} We use \textit{L2-norm} to evaluate the differences between values predicted by each of the models and the actual values. L2-norm represents the square root of the second sample moment of the differences between predicted values and observed values. For the predicted-vector $\hat{y}$ and actual output-vector $y$, \textit{L2-norm} is computed as follows in Eq. (\ref{eq:l2-norm}):
\begin{eqnarray}
\label{eq:l2-norm}
||\hat{y}-y||_2 = \left( \Sigma_{i=1}^{N} |\hat{y}-y|^{2}\right)
\end{eqnarray}

\begin{figure*}[!thb]
\includegraphics[width=0.9\textwidth]{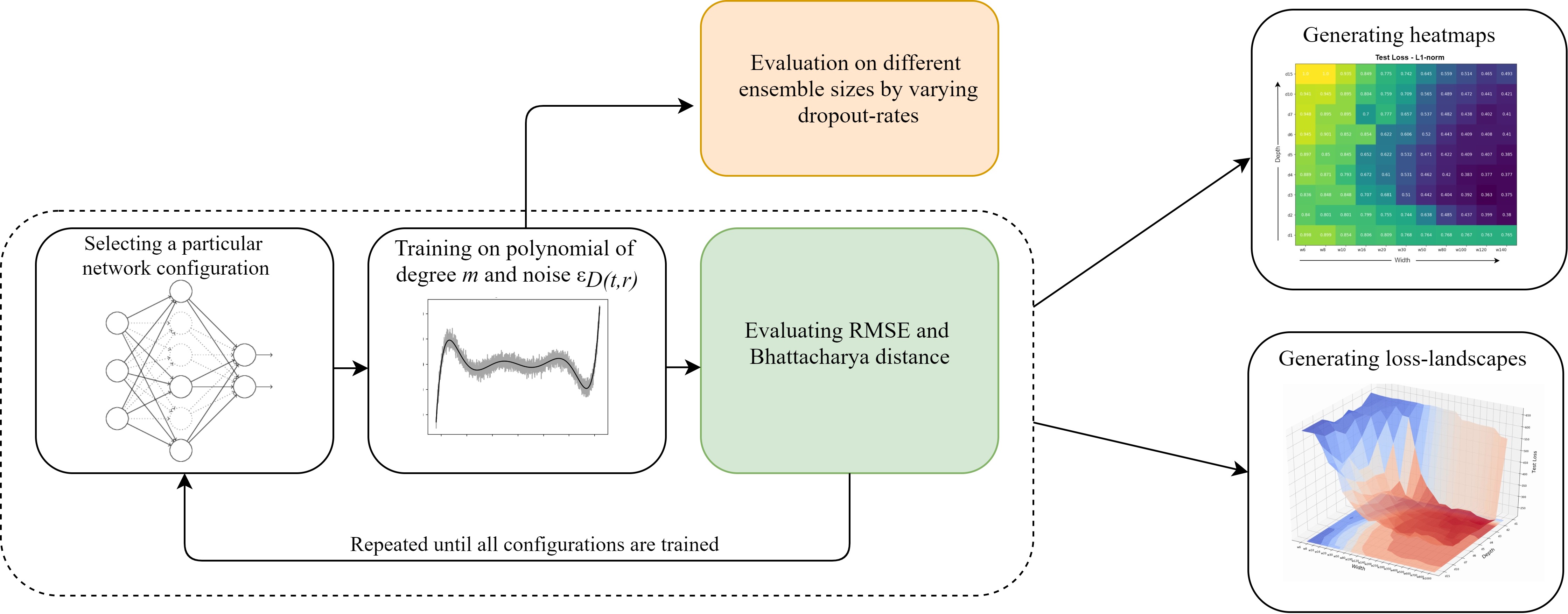}
\caption{\label{fig:pipeline} Experiment pipeline showing various steps involved to understand relationship between hyper-parameters and model performance for various network configurations. }
\end{figure*}

\textbf{Bhattacharyya distance:}  Bhattacharyya distance is one way of measuring the similarity between two probability distributions. In our case, we use \textit{Bhattacharyya distance} to measure the similarity between the noise distribution ($p_1$) present in the data and the residual ($p_2$) obtained form the prediction. \textit{Bhattacharyya distance} ($BD$) is calculated as follows in Eq. (\ref{eq:bd-dist}):

\begin{equation}
\begin{split}
\label{eq:bd-dist}
BD(p_1, p_2) = \frac{1}{4}ln\left(\frac{1}{4} \left(
\frac{\sigma_{p1}^{2}}{\sigma_{p2}^{2}} + \frac{\sigma_{p2}^{2}}{\sigma_{p1}^{2}} + 2\right)\right) \\+ \frac{1}{4} \left( \frac{(\mu_{p2} - \mu_{p1})^{2}}{\sigma_{p2}^{2} + \sigma_{p1}^{2}} \right)
\end{split}
\end{equation}

where $p_2 = \hat{y} - y$ and $p_1 = \epsilon_{D(t,r)}$ represents the noise from distribution $t$ and \textit{SNR}-level $r$.

While \textit{L2-norm} gives an estimation of how close the predictions are to the actual values, \textit{Bhattacharyya distance} between the residual and noise, on the other hand, gives an estimate of how much signal has been retained and what amount of bias has been induced on the network due to the added noise.





\section{\label{sec:results}Results and Analysis}
This section discusses the affect of varying the network size, the function complexity, and the amount of noise added to the training data on performance criteria including \textit{L2-norm} and \textit{Bhattacharyya distance}. \\



\begin{figure*}[!thb]
\includegraphics[width=\textwidth]{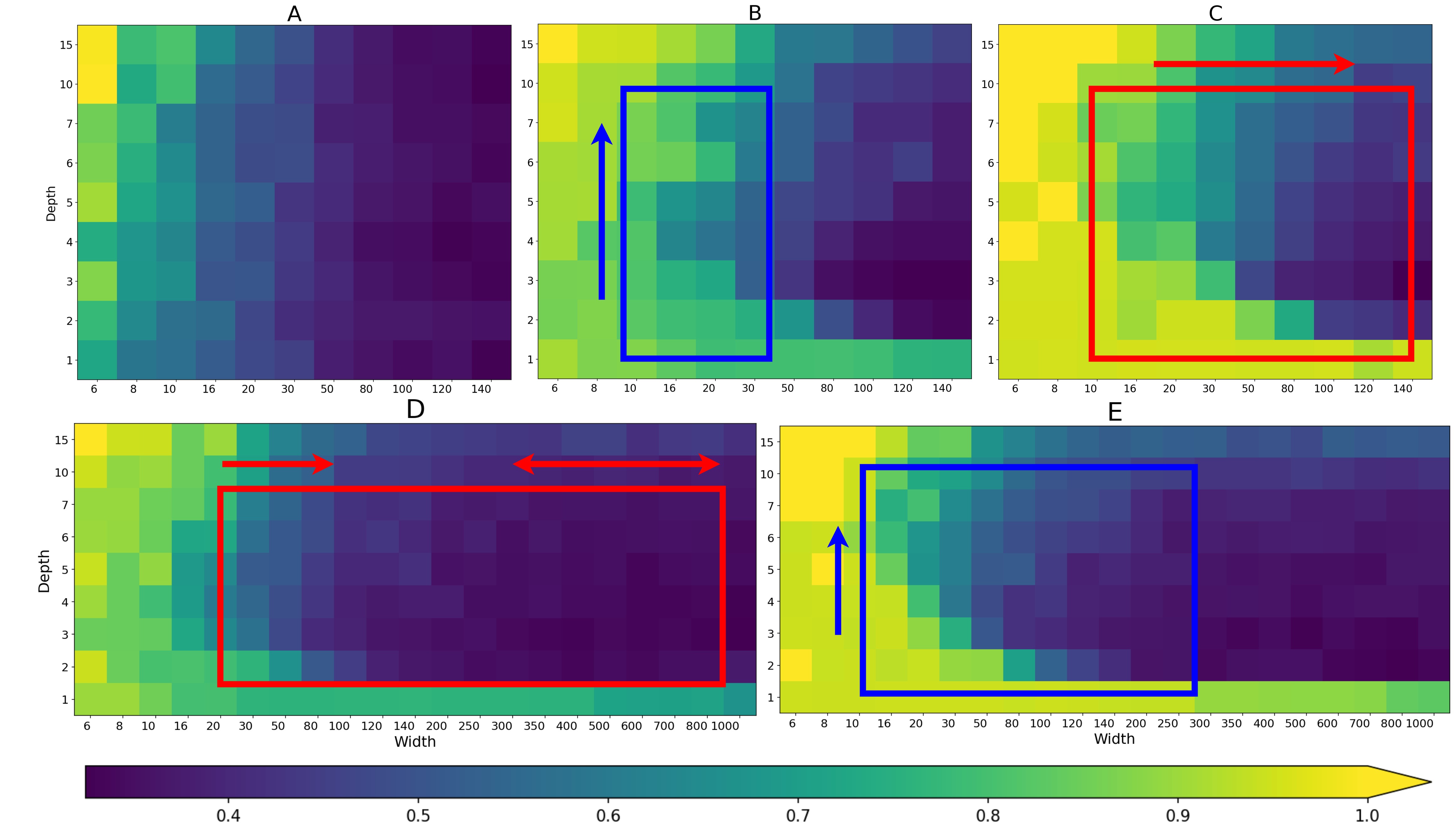}
\caption{\label{fig:heatmap2} Top row: Test Loss (\textit{L1-norm}) with increasing polynomial degree on Gaussian noise with $SNR=20$. Bottom row: Test Loss (\textit{L1-norm}) on networks with width ($w$) ranging from 6 to 1000.}
\end{figure*}

Figure~\ref{fig:heatmap2} (A-C) shows the test-set error (\textit{L1-norm}) as the number of hidden units (width \textit{w}) is varied from 6 to 140, and the depth ($d$) of the network from 1 to 15. We observe better performance in networks with larger width. 

\textbf{Increasing Width: }We increase the range of width ($w$) to 1000 to investigate the existence of "optimality" (used loosely in this research) in terms of width. Figure~\ref{fig:heatmap2} (D) shows the test loss \textit{L1-norm} on polynomial with degree 3 and right-side for degree 5. Both the polynomials are contaminated with \textit{Gaussian} noise having $SNR=20$. As we move in the direction of increasing width (red arrow), we notice constant decrements in test loss values. Also, notice the saturation of test loss on networks when increasing the width. We note the high level of test-set loss in case of 1-layer network. A possible explanation is that a single-layer is probably too small to accurately characterize the target function.

\textbf{Increasing Depth: }For each value of width, we train networks increase with depth values ranging from 1 to 15 to also assess the impact of depth parameter. From the right heatmap in Figure~\ref{fig:heatmap2} (E), as we move in the direction of increasing width (blue arrow), we observe decreasing values of test loss followed by a steady increase in test loss indicating deteriorating performance while incrementing after a certain depth-value in the feedforward neural network.\\ 


\begin{figure*}[!thb]
\includegraphics[width=1\textwidth]{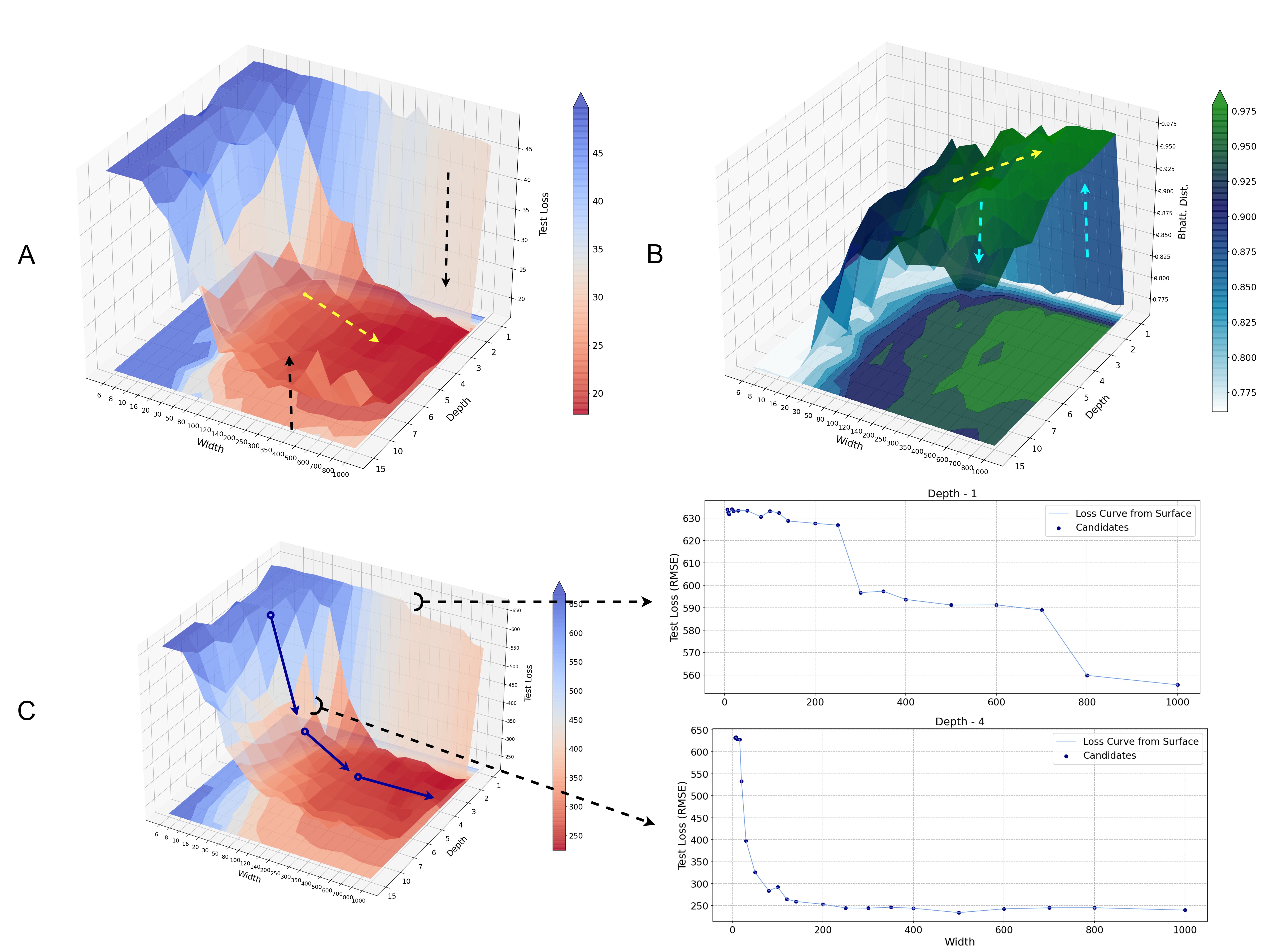}
\caption{\label{fig:loss_bd_landscape} A, B: Test Loss (\textit{L1-norm}) and Bhattacharya distance criterion landscapes on polynomial with degree 7 on exponential noise with $SNR=10$. C: Test Loss (\textit{L1-norm}) criterion landscape and individual loss breakdown along width on $5^{th}$ degree polynomial data with exponential noise and $SNR=10$.}
\end{figure*}

\subsection{Loss Landscapes}
Figure~\ref{fig:loss_bd_landscape} (A, B) shows the discussed loss landscapes of L1-norm (A) and Bhattacharya norm (B) criteria on out-of-distribution data points of the polynomial with degree 7 on exponential noise with $SNR=10$. If we consider the plot in Figure~\ref{fig:loss_bd_landscape} (A), it can be observed that there is a constant downward slope in the landscape as width increases (yellow arrow). On the other hand, when we consider models with higher depth, we first observe a steep downward slope followed by an upward slope leading to a valley-like structure. A corresponding one-to-one scenario is observed in landscapes based on Bhattacharya distance criterion. The only difference is that the surface becomes upside down compared to the curvature observed in the case of L1-loss. Figure~\ref{fig:loss_bd_landscape} (C) shows a surface breakdown of the L1-loss values along with the depth values, which allows for a more apparent observation of decreasing L-1 test values of candidates with higher width values.



Based on the results obtained, it can be inferred that there exists no optimality in terms of width in neural network on both L-1 loss and Bhattacharya distance criterion. Since, an increment in performance for the task of polynomial approximation is observed on models with higher width, therefore, a higher width is favourable. However, we do see that both L1-loss values and Bhattacharya distance criterion values starts saturating, we can consider it a performance gain vs model complexity tradeoff along the width. Therefore, on the basis of a fixed upper-threshold of model complexity or computational overheads, considering a model with maximum permissible width value is expected to perform better than models with lower width values given other hyperparameters are fixed.

While the analysis of candidates along increasing depth showed an increase followed by decrease in performance values in both the criterions indicating the existence of an optimal value of depth located in the region of changing slope direction. However, this depth value is not universal across all the width values, which means that the exact value of the optimal depth depends on the value of width chosen. Mathematically, we can say that, for a given width $w’ \exists$ depth $d’$ such that $d’$ is optimal for all possible $G(d, w’, m), d \ge 1$\\

\subsection{Effect of Ensembles}

Using the Monte-Carlo dropout method as a Bayesian approximation~\cite{gal2016dropout}, we consider our analysis on the third parameter, that is, the number of ensembles. By allowing dropouts in the test time, we conduct multiple inferences and consider the average of all the obtained models as the final output. Furthermore, the dropout strategy ensures that each time a different model is obtained with high probability, which helps in bringing diversity and thus reducing variance in the final model output.
Mathematically, this can be shown as:

\begin{eqnarray}
\hat{y} = \frac{1}{N}\Sigma_{i}^{m}f_i{x}
\end{eqnarray}

\begin{equation}
\begin{split}
\label{eq:ensemble}
E^{2} = (y- \hat{y})^2 = \frac{1}{N}\Sigma_{i}^{m}(y- f_i(x))^2 \\- \frac{1}{N}\Sigma_{i}^{m}(f_i(x)-\hat{y})^2
\end{split}
\end{equation}

where $\hat{y}$, $y$ and $f_i$ represents the final model output,  the ground truth  and the $i^{th}$ ensemble member respectively. The breakdown of $(y- \hat{y})^2$ in two parts shows why ensembles work better than a single model. The first term is the actual error whereas second term represents the disagreement between ensembles. \\

In comparison to the Mean-squared-error (MSE) for a single model, ensembles introduces another term (MSE between ensembles and the model output) as shown in Equation~\ref{eq:ensemble} which contributes in reducing the overall MSE of the model~\cite{hansen1990neural}.  


\begin{figure}[!thb]
\includegraphics[width=0.5\textwidth]{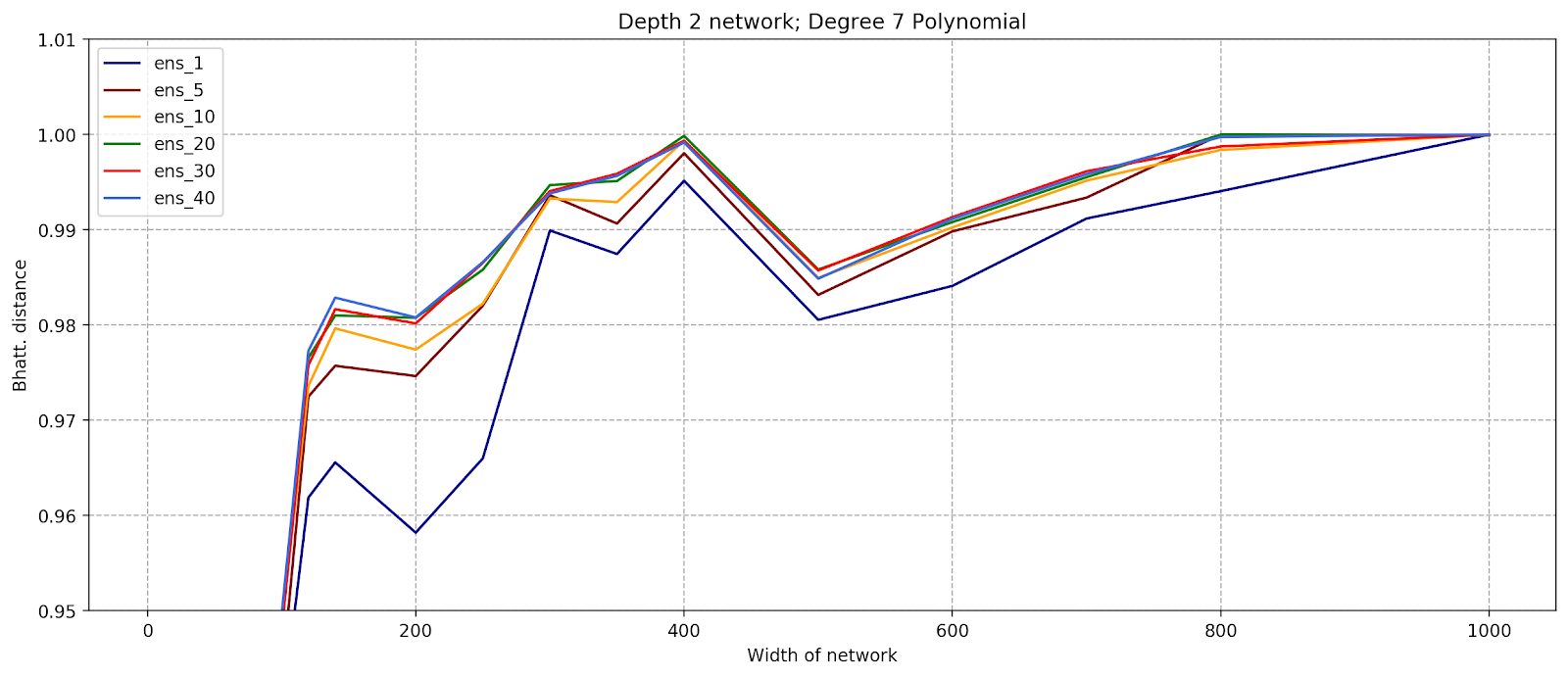}
\caption{\label{fig:ensembles} Bhattacharya criterion values of different ensemble sozes with increasing network width on 7 degree polynomial and Gaussian noise with $SNR=20$.}
\end{figure}

Figure~\ref{fig:ensembles} shows the Bhattacharya criterion values of models with different number of ensembles. A sudden increase in performance is observed as we go from a single model to an ensemble. But the performance gain on further increasing the ensembles size is observed to decrease. The reason for this observation is that the chances of finding uncorrelated models as we increase the number of ensembles decreases. If a particular ensemble size already incorporates the top-performing models then further adding  other members will not offer any benefit to the ensemble. From Figure~\ref{fig:ensembles}, it can be seen that in many networks, the highest ensemble size is not the best performer. Therefore, an optimal ensemble size may exist beyond which an improvement in performance isn't expected as the ensemble will not able to be harness their contribution effectively.\\       

\section{\label{sec:conc}Conclusion and Discussion}

We observed interesting relations between the choice of BDL hyperparameters and corresponding skill metrics for predictions with uncertainty. While directly relevant for the approximation of noisy polynomials, the insights may be directional to explore BDL-based robust function approximation in wider settings. 

The surface curvature obtained in our results showed that as depth increases past an optimal point, generalization performance tends to decrease. This experimental determination of the existence of an optimal depth value in polynomial function approximation tasks can allow for reduced number of trials needed to find the best BDL model in practical settings.

The observation of a continuously decreasing but positive performance gain with increasing width in our results indicated that an optimal width may not exist. However, the gain in generalization performance beyond a certain high value of width were observed to grow less statistically significant. These set of results suggest that the width may be chosen based on a threshold related to model complexity (e.g., a modified information criteria where model complexity and performance gain on out-of-sample data may need to be balanced) or based on the available computational resources.

We observed, through distributional distance metrics, that an optimal is suggested for ensemble sizes in a dropout-based BDL. In other words, the best approximations to the noise statistics (used to contaminate the underlying polynomials) were obtained for certain optimal ensemble sizes. These set of insights may be useful in the practical design and implementation of BDLs.

Our results point to the existence of an optimal depth and an optimal ensemble size but no optimal width for BDL representations. The empirical insights presented here may benefit from a closer relation to existing and potentially new learning theory in these areas. Future experiments on simulated data need to examine a broader set of simulations, including different types of (potentially nonlinear) signals as well as (potentially correlated) noise processes, in both time and space. Practical BDL guidelines must be developed across multiple science, engineering, and business domains, by considering use cases based on both realistic simulations as well as on real data.
\nocite{*}

\begin{acknowledgments}
ARG was supported by four National Science Foundation Projects including NSF BIG DATA under Grant No. 1447587, NSF Expedition in Computing under Grant No. 1029711, NSF CyberSEES under Grant No. 1442728, and NSF CRISP type II under Grant No. 1735505. UB was supported by the Ministry of Education (India), STARS under project Project ID: 367. ARG acknowledges his guest affiliation at IIT Gandhinagar which made this work feasible. 
\end{acknowledgments}

\bibliography{apssamp}

\end{document}